\documentclass{article}

    \PassOptionsToPackage{numbers, compress}{natbib}

     \usepackage[final]{neurips_2019}

\usepackage[utf8]{inputenc} 
\usepackage[T1]{fontenc}    
\usepackage{hyperref}       
\usepackage{url}            
\usepackage{booktabs}       
\usepackage{amsfonts}       
\usepackage{nicefrac}       
\usepackage{microtype}      

\usepackage{multirow}

\usepackage{amssymb}
\usepackage{amsmath}
\usepackage{graphicx}

\usepackage{bm}
\newcommand{\vect}[1]{\bm{#1}}
\newcommand{\matr}[1]{\bm{#1}}

\newcommand{\vh}[0]{\vect{h}}

\newcommand{\vx}[0]{\vect{x}}

\newcommand{\vf}[0]{\vect{f}}

\newcommand{\vo}[0]{\vect{o}}
\newcommand{\vy}[0]{\vect{y}}
\newcommand{\vg}[0]{\vect{g}}

\newcommand{\mW}[0]{\matr{W}}

\usepackage{url}

\title{Modulated Self-attention Convolutional Network for VQA}

\author{
  Jean-Benoit Delbrouck$^1$
  \And
  Antoine Maiorca$^1$
  \And
   Nathan Hubens$^{1,2}$
  \And
   St\'ephane Dupont$^1$
  \and
  $^1$ ISIA Lab, Polytechnic Mons, Belgium \\
  \and
  $^2$ Telecom SudParis, France \\
  \and
  \{jean-benoit.delbrouck, antoine.maiorca, nathan.hubens, stephane.dupont\}@umons.ac.be
  \\
}

\begin{document}

\maketitle

\begin{abstract}
As new data-sets for real-world visual reasoning and compositional question answering are emerging, it might be needed to use the visual feature extraction as a end-to-end process during training. This small contribution aims to suggest new ideas to improve the visual processing of traditional convolutional network for visual question answering (VQA). In this paper, we propose to modulate by a linguistic input a CNN augmented with self-attention. We show encouraging relative improvements for future research in this direction.
\end{abstract}

\section{Introduction}
Problems combining vision and natural language processing such as visual question answering (VQA) is viewed as an extremely challenging task. Visual attention-based neural decoder models \cite{pmlr-v37-xuc15,KarpathyL15} have been widely adopted to solve such tasks. All recent works pushing the state-of-the-art in VQA are using the so-called bottom-up attention features \cite{anderson2018bottom}. It consists of multiple pre-extracted features corresponding to local regions of interest in an image. Therefore, it is not  possible to modulate the entire visual 
pipeline to capture relationships or make comparisons between widely separated spatial regions according to the question. For example, questions of the new GQA data-set \cite{hudson2018gqa} tackles reasoning skills such as object and attribute recognition, transitive relation tracking, spatial reasoning, logical inference and comparisons. This contribution aims to bring back the features extraction process end-to-end in the learning and to modulate the convolutional network by a linguistic input though self-attention.

\section{Related work}

Two main approaches have helped pushing further the state-of-the-art in Visual Question Answering: the co-attention module and the bottom up image features. Our work also takes inspiration in the recent attention augmented Convolutional Networks.

\textbf{Image features} \quad Most early results have used a VGG \cite{VQA} or a ResNet \cite{lu2016hierarchical} CNN pretrained on ImageNet \cite{russakovsky2015imagenet} to extract visual features from images. Recently, richer bottom-up attention features \cite{anderson2018bottom} have been proposed. It consists of pre-extracted from a Faster R-CNN (that outputs bounding boxes of interest) in conjunction with ResNet-101 (that acts as a feature extractor for each selected bounding box). They are widely used in the most recent works as they have shown great success for VQA and image captioning. 

\textbf{Co-attention} \quad Attention networks in multimodal learning provide an efficient way to utilize given
visual information selectively. However, the computational cost to learn attention
distributions for every pair of multimodal input channels is prohibitively expensive.
To solve this problem, a co-attention learning framework is proposed  \cite{chen2017deeplab} to reduce the expensive computational cost to learn attention distributions for every pair of multimodal input channels. An improvement to the co-attention method is introduced by \cite{yu2018beyond} that consists of two steps, a self-attention for a question embedding and the question-conditioned attention for a visual embedding. However, these co-attention approaches use separate attention distributions for each modality, neglecting the interaction between the modalities what we consider and model. To tackle this problem, \cite{kim2018bilinear} propose bilinear attention networks that find bilinear attention distributions to effectively utilize vision-language information. Finally, \cite{yu2019deep} propose a Deep Modular Co-Attention Networks composed of multiple blocks of two basic attention units : a self-attention unit (inspired by machine translation \cite{vaswani2017attention}) and the guided-attention (GA) unit to model the dense inter-modal interactions. The authors also concatenate the res5c features from ResNet-152 to the bottom up features from \cite{anderson2018bottom}.

\textbf{Attention augmented CNN} \quad Over the years, multiple attention mechanisms have been proposed for visual tasks to address the weaknesses of convolutions. For instance, Squeeze-and-Excitation networks \cite{hu2018squeeze} reweight feature channels using signals aggregated from entire feature maps while CBAM \cite{woo2018cbam} module sequentially infers attention maps along two separate the channel and spatial independently. More recent attention augmented networks produces additional feature maps by using Attention Augmented Convolution  \cite{bello2019attention} or recalibrates the feature maps via addition \cite{zhang2019self}.

\section{Attention augmented Residual Network}

\subsection{Residual Network} \label{resnet}
ResNets \cite{he2016deep} are built from residual blocks:

$$\vy = \mathcal{F}(\vx, \{\mW_i\}) + \mW_s\vx $$

Here, $\vx$ and $\vy$ are the input and output vectors of the layers considered. The function $\mathcal{F}(\vx, \{\mW_i\})$ is the residual mapping to be learned. For an example, if we consider two layers, $\mathcal{F} = \mW_2 \sigma ( \mW_1 \vx)$ where $\sigma$ denotes ReLu function. The operation $\mathcal{F} + \vx$ is the shortcut connection and consists of an element-wise addition. Therefore, the dimensions of $\vx$ and $\mathcal{F}$ must be equal. When this is not the case (e.g., when changing the input/output
channels), the $\mW_s$ matrix performs a linear projection by the shortcut connections to match the dimension. Finally, it performs a last second nonlinearity after the addition (i.e., $\sigma(y))$. A group of blocks are stacked to form a stage of computation. The general ResNet architecture starts with a single convolutional layer followed by 4 stages of computation.

\subsection{Self-attention layer}

A self-attention layer can be placed in-between two layers in a residual block. Let's consider the image features from the previous layer: $\boldsymbol{x} \in \mathbb{R}^{C \times N}$. Here, $C$ is the number of channels and $N$ is the number of feature locations of features.

Features $x$ are first transformed into two feature spaces $ \boldsymbol{f}, \boldsymbol{g}$, where $\boldsymbol{f}(\boldsymbol{x})=\boldsymbol{W}_{f} \boldsymbol{x}, \boldsymbol{g}(\boldsymbol{x})=\boldsymbol{W}_{g} \boldsymbol{x}$. We use $\vf$ and $\vg$ to compute the square matrix $N\times N$ of attention weights $\beta$:

\begin{equation}
\beta_{j, i}=\frac{\exp \left(s_{i j}\right)}{\sum_{i=1}^{N} \exp \left(s_{i j}\right)}, \text{ where } s_{i j}=\boldsymbol{f}\left(\boldsymbol{x}_{i}\right)^{T} \boldsymbol{g}\left(\boldsymbol{x}_{j}\right)
\end{equation}

With $\beta_{j, i}$ indicating the extent to which the model attends to
the $i^{t h}$ location when synthesizing the $j^{t h}$ location. The output of the attention layer is $\boldsymbol{o}=\left(\boldsymbol{o}_{1}, \boldsymbol{o}_{2}, \ldots, \boldsymbol{o}_{j}, \ldots, \boldsymbol{o}_{N}\right) \in$
$\mathbb{R}^{N \times C},$ where

\begin{equation} \label{equationO}
\boldsymbol{o}_{j}=\sum_{i=1}^{N} \beta_{j, i} \boldsymbol{k}\left(\boldsymbol{x}_{i}\right), \boldsymbol{k}\left(\boldsymbol{x}_{i}\right)=\boldsymbol{W}_{\boldsymbol{k}} \boldsymbol{x}_{i}
\end{equation}

In the above formulation, $\boldsymbol{W}_{\boldsymbol{g}} \in \mathbb{R}^{C \times \overline{C}}, \boldsymbol{W}_{\boldsymbol{f}} \in \mathbb{R}^{C \times \overline{C}}$
$\boldsymbol{W}_{\boldsymbol{k}} \in \mathbb{R}^{C \times C}$are the learned weight matrices, which are implemented as $1 \times 1$ convolutions. \\

Additionally, the output is multiplied by a factor $\gamma$ and we add back the input feature map.

\begin{equation} \label{gamma_origin}
\boldsymbol{y}_{i}=\gamma \boldsymbol{o}_{i}+\boldsymbol{x}_{i}
\end{equation}

$\gamma$ is a learnable scalar and it is initialized as $0$. Introducing the learnable $\gamma$ allows the visual network to first rely on the cues in the local neighborhood and the language model to converge normally. As the gamma goes up, the model will gradually learn long range interactions as required for a task such as VQA.

\section{Linguistic modulation}

In this section, we described two methods to modulate a pretrained ResNet through the self attention modules. We want to enable the linguistic embedding to manipulate the self-attention mechanism. We decide to do so through the parameter $\gamma$ and the attention weights $\beta$. 

$\boldsymbol{\gamma}$ \textbf{modulation} \quad Given the last hidden state $\vh_T$ of the RNN encoding the question, we output a modulated $\gamma^{\text{m}}$ given by :
\begin{equation} \label{gammamod}
\gamma^h = \mW_h \vh_T
\end{equation}
where $\mW_h \in \mathbb{R}^{|\vh_T| \times 1}$. Recall that in the previous section, $\gamma$ in equation \ref{gamma_origin} is unique for any training examples. Here, we output a dedicated scalar for every example in the batch. We rewrite equation \ref{gamma_origin} as :

\begin{equation} 
\boldsymbol{y}_{i}=\gamma^h_{i} \boldsymbol{o}_{i}+\boldsymbol{x}_{i}
\end{equation}

$\boldsymbol{\beta}$ \textbf{modulation}

We define new linguistic and visual features spaces $ \boldsymbol{p}, \boldsymbol{q}$ where $\boldsymbol{p}(\boldsymbol{h})=\boldsymbol{W}_{p} \boldsymbol{h}$ and $\boldsymbol{q}(\boldsymbol{x})=\boldsymbol{W}_{q} \boldsymbol{x}$. Both feature spaces are used to compute a new set of attention weights $\beta^h$ in the following manner:

\begin{equation}
\beta^h_{j, i}=\frac{\exp \left(s_{i j}\right)}{\sum_{i=1}^{N} \exp \left(s_{i j}\right)}, \text{ where } s_{i j}=\boldsymbol{q}\left(\boldsymbol{x}_{i}\right)^{T} \boldsymbol{p}\left(\boldsymbol{h}_{j}\right)
\end{equation}

Where $\beta^h_{j, i}$ indicating the extent to which the model attends to
the $i^{t h}$ spatial location of feature map $\vx$ when synthesizing the $j^{t h}$ dimension of the hidden state. The hidden state $h_T$ is a vector therefore $j=1$ and $\beta^h \in \mathbb{R}^{N\times1}$. We apply this set of betas to the output $\vo$ defined in equation \ref{equationO}. This modulation can be seen as a linguistic spatial attention on the visual self attention: each column $\beta_{j}$ is re-weighted by the scalar $\beta^h_{j}$.

\section{Experiments}
\textbf{Settings} \quad We use the VQA v2.0 train and validation \cite{balanced_vqa_v2} consisting of 443,757 and 214,354 questions respectively over 123,287 images. As VQA model, we follow the Bottom-Up and Top-Down Attention from \cite{anderson2018bottom}. We replace the Up-Down features with a ResNet as visual features extractor (ResNet-34 for preliminary experiments and a ResNet-152 \cite{he2016deep} for final results). The ResNet is pretrained on ImageNet \cite{russakovsky2015imagenet} and frozen during training. Only the self-attention weights are training parameters. For each images, we extract features at the end of the third ResNet stage. \\

Learning parameters are trained with Adamax optimizer and a learning rate of $2e^{-2}$. In a block, the self-attention module is always placed between the last convolution and batch normalization layer of the said block.

\textbf{Block and layer search} \quad 
Thanks to the attention, the model can check that features in distant portions of the image are consistent with each other.
Depending on where we place one or several attention modules, the model can compute affinity scores between low-level or high-level features in distant portions of the image. It is also worthy to note that attention module in early stages of ResNet is computationally expensive (i.e. the first stage has $N = 3136$ spatial locations).

\begin{table}[!htb]
	\label{sample-table}
    \begin{minipage}{.400\textwidth}
            \begin{tabular}{llllllll}
            \multicolumn{1}{c}{\bf ResNet-34}  &
            \multicolumn{1}{c}{\textbf{Eval set \%}}&
            \multicolumn{1}{c}{\textbf{\#param}}\\
            \\ \hline \\
            Baseline (No SA)\cite{anderson2018bottom} & 55.00 & 0M  \\
            SA (S: 1,2,3 - B: 1)  & 55.11 &   \multirow{3}{*}{ \Bigg{\}} 0.107M} \\
            SA (S: 1,2,3 - B: 2)  & 55.17 \\
            \textbf{SA (S: 1,2,3 - B: 3)} & \textbf{55.27} \\
            \hline \\  \\  \\  \\
\end{tabular}
    \end{minipage}%
      \hspace{15mm}
    \begin{minipage}{.400\textwidth}
      \centering
             \vspace{0pt}

\begin{tabular}{llllllll}
            \multicolumn{1}{c}{\bf ResNet-34}  &
            \multicolumn{1}{c}{\textbf{Eval set \%}}   &
            \multicolumn{1}{c}{\textbf{\#param}}\\
            \\ \hline \\
            SA (S: 3 - M: 1) & 55.25 & \multirow{4}{*}{\Bigg{\}} 0.082M} \\
            \textbf{SA (S: 3 - B: 3)} & \textbf{55.42} \\
            SA (S: 3 - B: 4) & 55.33 \\
            SA (S: 3 - B: 6) & 55.31 \\
            \hline \\
            SA (S: 3 - B: 1,3,5) & 55.45 & \multirow{2}{*}{\Big{\}} 0.245M} \\
            \textbf{SA (S: 3 - B: 2,4,6)} & \textbf{55.56} \\ \\
\end{tabular}

    \end{minipage} 
              \caption{Experiments run on a ResNet-34. Numbers following S (stages) and B (blocks) indicate where SA (self-attention) modules are put. Parameters count concerns only SA and are in millions (M).}

\end{table}

We empirically found that self-attention was the most efficient in the 3rd stage. It is also the less computationally expensive \footnote{Early stages requires less training parameters as $C$ is smaller, but $f^Tg$ computation is more expensive}. It is also beneficial to focus the last block of a stage rather than the first blocks. We notice small improvements relative to the baseline showing that self-attention alone does improve the VQA task.

\textbf{Linguistic modulation} \quad We noticed that the current architecture was not able to learn the $\boldsymbol{\gamma}$ \textbf{modulation} properly: equation \ref{gammamod} would most of the time result in a large scalar or close to 1. when a sigmoid layer is added. Therefore, the self-attention weight is already too important at the beginning, skipping visual local connections are misleading the overall model training. However, we managed to show improvements with the $\boldsymbol{\beta}$ \textbf{modulation} with a ResNet-152. Though the improvement is slim, it is encouraging to continue researching into visual modulation

\begin{table}[!htb]
	\label{sample-table}
    \begin{minipage}{.400\textwidth}
            \begin{tabular}{llllllll}
            \multicolumn{1}{c}{\bf ResNet-34}  &
            \multicolumn{1}{c}{\textbf{Eval set \%}}&
            \multicolumn{1}{c}{\textbf{\#param}}\\
            \\ \hline \\
            SA (S: 3 - B: 1,3,5) & \textbf{55.56} & 0.245M \\
            + $\boldsymbol{\gamma}$ \textbf{modulation}  & 52.26  & 0.248M \\
            + $\boldsymbol{\beta}$ \textbf{modulation}  & 55.32 & 1.573M  \\
             \\ \\
\end{tabular}
    \end{minipage}%
      \hspace{15mm}
    \begin{minipage}{.400\textwidth}
      \centering
             \vspace{0pt}

\begin{tabular}{llllllll}
            \multicolumn{1}{c}{\bf ResNet-152}  &
            \multicolumn{1}{c}{\textbf{Eval set \%}}   &
            \multicolumn{1}{c}{\textbf{\#param}}\\
            \\ \hline \\
            Baseline (No SA)\cite{anderson2018bottom} & 57.10 & 0M 
            \\ \hline \\
            SA (S: 3 - B: 2,18,36) & 58.05 & 3.932M \\
            + $\boldsymbol{\beta}$ \textbf{modulation}  & \textbf{58.35} & 15.731M  \\ \\
\end{tabular}

    \end{minipage} 
              \caption{Experiments run on a ResNet-34 and 152 for the $\gamma$ and $\beta$ modulation.}

\end{table}

It is worthy to note that the ResNet baseline of \cite{anderson2018bottom} reaches a 57.9\% accuracy and was achieved with a ResNet-200. Our best model achieves 58.35\% with a ResNet-152.

\section{Conclusion}

We showed that it is possible to improve the feature extraction procedure for the VQA task by adding self-attention modules in the different ResNet blocks. The improvement margins can be further improved by:

\begin{itemize}
    \item \textbf{Using wider CNN}: Now that we include the feature extraction process end-to-end in the learning, we could pick wider CNN such as ResNeXt \cite{xie2017aggregated} or Wide-ResNet \cite{zagoruyko2016wide};
    \item \textbf{Better self-attention}: More sophisticated self-attention, such as multi-head attention in CNN \cite{bello2019attention, yu2019deep} could improve the overall model;
    \item \textbf{Better modulation}: The linguistic hidden state should directly be an input of the computation for self-attention. Other works have already considered using all the hidden states from the language model as the "key" $K$ for visual self-attention \cite{yu2019deep}.
    \item \textbf{Other modulation}: The self-attention modulation could be coupled with modulated batch-normalization as previously investigated in \cite{de2017modulating}.
\end{itemize}

\bibliographystyle{unsrt}
\bibliography{nips2019}
\end{document}